\documentclass[sigconf]{aamas}

\usepackage{balance} % for balancing columns on the final page
\usepackage{xcolor}
\usepackage{hyperref}
\hypersetup{
    colorlinks=true,
    linkcolor=blue,
    urlcolor=blue,
    citecolor=red
}
\usepackage{cleveref}
\usepackage{multicol}
\usepackage{algorithm}
\usepackage{algpseudocode}
\usepackage{subcaption} 
\usepackage{dblfloatfix}
\usepackage{enumitem}

\makeatletter
\gdef\@copyrightpermission{
  \begin{minipage}{0.2\columnwidth}
   \href{https://creativecommons.org/licenses/by/4.0/}{\includegraphics[width=0.90\textwidth]{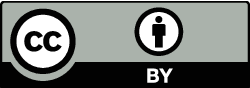}}
  \end{minipage}\hfill
  \begin{minipage}{0.8\columnwidth}
   \href{https://creativecommons.org/licenses/by/4.0/}{This work is licensed under a Creative Commons Attribution International 4.0 License.}
  \end{minipage}
  \vspace{5pt}
}
\makeatother

\setcopyright{ifaamas}
\acmConference[AAMAS '25]{Proc.\@ of the 24th International Conference
on Autonomous Agents and Multiagent Systems (AAMAS 2025)}{May 19 -- 23, 2025}
{Detroit, Michigan, USA}{Y.~Vorobeychik, S.~Das, A.~Nowé  (eds.)}
\copyrightyear{2025}
\acmYear{2025}
\acmDOI{}
\acmPrice{}
\acmISBN{}

\acmSubmissionID{<<OpenReview submission id>>}

\title[AAMAS-2025 MHRI]{Human-Agent Coordination in Games under Incomplete Information via Multi-Step Intent}

\author{Shenghui Chen}
\authornote{Equal contribution.}
\affiliation{
  \institution{University of Texas at Austin}
  \city{Austin}
  \country{United States}}
\email{shenghui.chen@utexas.edu}

\author{Ruihan Zhao}
\authornotemark[1]  % Reuses the previous authornote
\affiliation{
  \institution{University of Texas at Austin}
  \city{Austin}
  \country{United States}}
\email{ruihan.zhao@utexas.edu}

\author{Sandeep Chinchali}
\affiliation{
  \institution{The University of Texas at Austin}
  \city{Austin}
  \country{United States}}
\email{sandeepc@utexas.edu}

\author{Ufuk Topcu}
\affiliation{
  \institution{The University of Texas at Austin}
  \city{Austin}
  \country{United States}}
\email{utopcu@utexas.edu}

\begin{abstract}
Strategic coordination between autonomous agents and human partners under incomplete information can be modeled as turn-based cooperative games. 
We extend a turn-based game under incomplete information, the shared-control game, to allow players to take multiple actions per turn rather than a single action. The extension enables the use of multi-step intent, which we hypothesize will improve performance in long-horizon tasks.
To synthesize cooperative policies for the agent in this extended game, we propose an approach featuring a memory module for a running probabilistic belief of the environment dynamics and an online planning algorithm called \textsc{IntentMCTS}. This algorithm strategically selects the next action by leveraging any communicated multi-step intent via reward augmentation while considering the current belief.
Agent-to-agent simulations in the Gnomes at Night testbed demonstrate that \textsc{IntentMCTS} requires fewer steps and control switches than baseline methods. A human-agent user study corroborates these findings, showing an $18.52\% $ higher success rate compared to the heuristic baseline and a $5.56\%$ improvement over the single-step prior work. 
Participants report lower cognitive load, frustration, and higher satisfaction with the \textsc{IntentMCTS} agent partner. \footnote{Project page: \url{https://vivianchen98.github.io/mhri_intent_aamas_page/}.}
\end{abstract}


\newcommand{\BibTeX}{\rm B\kern-.05em{\sc i\kern-.025em b}\kern-.08em\TeX}
\newcommand{\Xcal}{\mathcal{X}}
\newcommand{\Scal}{\mathcal{S}}
\newcommand{\Acal}{\mathcal{A}}
\newcommand{\Tcal}{\mathcal{T}}
\newcommand{\Rcal}{\mathcal{R}}

\newcommand{\human}{\mathit{H}}
\newcommand{\ego}{\mathit{E}}

\definecolor{egoblue}{HTML}{4472C4} % Defining a color using hex code
\definecolor{humangreen}{HTML}{70AD47}

\definecolor{humanhumangray}{HTML}{B7B7B7}
\definecolor{agentagentorange}{HTML}{F2A95D}
\definecolor{agentmutebrown}{HTML}{B88A6D}

\definecolor{pptblue}{HTML}{4472C4}

\theoremstyle{definition}

\newtheorem{theorem}{Theorem}
\newtheorem{definition}{Definition}
\newtheorem{problem}{Problem}

\newcommand{\argmax}[1]{\operatorname{arg\,max}\limits_{#1}}

\DeclareMathOperator{\subjectto}{s.t.}

\begin{document}

%%% The following commands remove the headers in your paper. For final 
%%% papers, these will be inserted during the pagination process.

\pagestyle{fancy}
\fancyhead{}

%%% The next command prints the information defined in the preamble.

\maketitle 

%%%%%%%%%%%%%%%%%%%%%%%%%%%%%%%%%%%%%%%%%%%%%%%%%%%%%%%%%%%%%%%%%%%%%%%%

\section{Introduction}\label{sec:introduction}
% goal: autonomous agent capable of strategic coordination with human partners under incomplete information in long-horizon tasks
% importance: potential for deliberate strategizing partner in various domains
Developing autonomous agents that can strategically coordinate with human partners under incomplete information, especially in long-horizon tasks, is important in the field of human-agent interaction. These agents hold the potential to improve interaction across various contexts, ranging from virtual applications like strategy board games \cite{bard2020hanabi} and action video games \cite{carroll2019utility} to physical domains like assistive technologies such as wheelchairs \cite{goil2013using}.

% challenge: multi-step planning and strategizing
A key challenge in achieving human-agent coordination is multi-step planning and strategizing. Humans naturally rely on multi-step planning \cite{miller2021multi} and frequently engage in multi-step intent communication, essential for deliberate coordination in long-horizon tasks.
Therefore, for autonomous agents to be effective collaborators, they should also understand and perform multi-step reasoning.

% challenge: prior work only copes with single-step intents, cannot capture the richness in strategy and does not do well in long-horizon tasks; also it deals with 

% current gap: existing models are not good enough
% While models like partially observable Markov decision processes (POMDPs) and partially observable stochastic games (POSGs) address imperfect information in single-agent and multi-agent settings, they primarily focus on partial observability of states, not dynamics such as transition functions. 
An existing model for human-agent cooperation under incomplete information, the \textit{shared-control game} introduced in \cite{chen2024human}, studies scenarios where each player has access only to its own transition function but not its partner's. However, this model is limited to a single action per turn, preventing the realization of more intricate, multi-step strategies. Although recent work in \cite{chen2024learning} partially mitigates this issue by allowing one of the players to take multiple actions in its turn, a full extension that supports multi-action dynamics for both players remains unexplored.

\begin{figure}[t]
    \centering
    \includegraphics[width=0.45\linewidth]{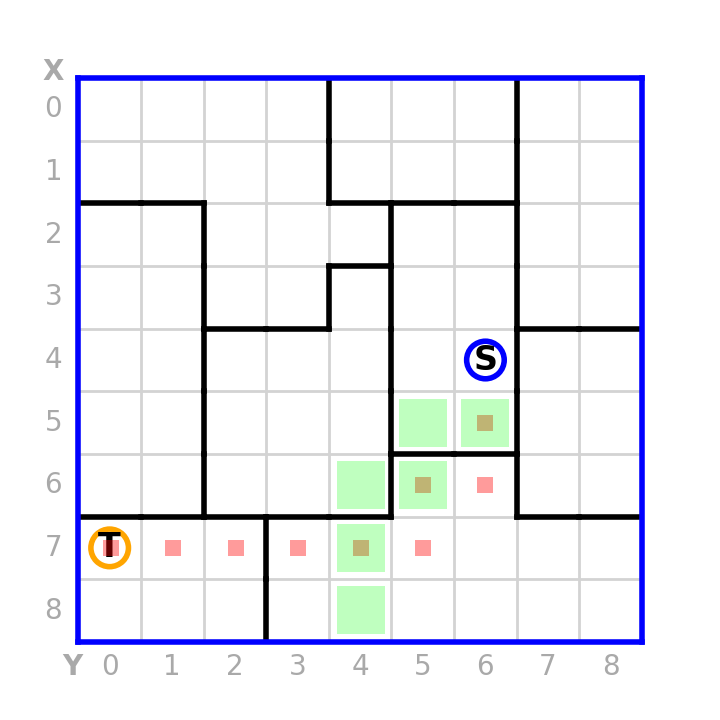}
    \raisebox{0.07cm}{\includegraphics[width=0.424\linewidth]{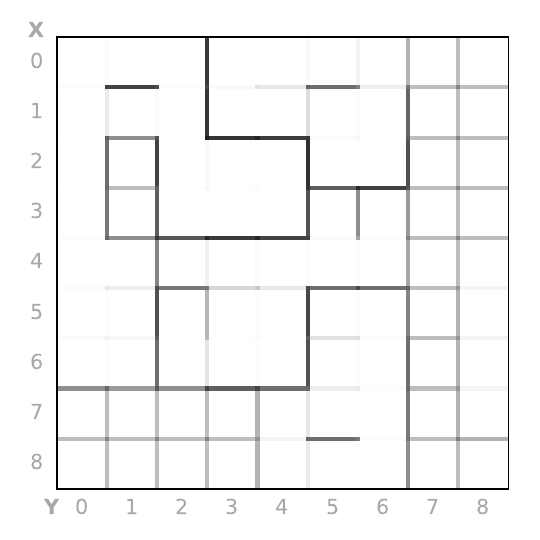}}
    \raisebox{0.08cm}{\includegraphics[trim={9.15cm 0 0 0},clip,height=3.56cm]{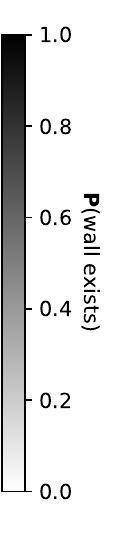}}
    \vspace{-1.5em}
    \caption{Snapshot of Gnomes at Night gameplay from the agent's perspective. \textit{Left}: The agent's side of the maze, with the human's intent marked in large green squares and its own intent in small red squares. \textit{Right}: The agent’s probabilistic belief of the wall layout in the human’s maze.}
    \label{fig:cover}
    \vspace{-1em}
\end{figure}

% model
We extend the shared-control game by allowing both players to take multiple actions per turn, capturing more fluid interactions and bringing the model closer to natural human-agent coordination. This extension enables the agent to leverage \textit{multi-step intent}, which was previously unattainable due to the limitation of single-action turns. In this work, we define a multi-step intent as a variable-length sequence of states a player wishes its partner would visit next.
% problem
% In this modified game model, we pose the problem of intent-aware cooperative policy synthesis, where an agent aims to maximize the discounted total reward while adhering to the game transitions by effectively coordinating with a partner through exchanging multi-step intents and observing their transition histories.
The goal is to develop an agent, as the ego player in this game, that infers partner dynamics through their transition histories and coordinates effectively by exchanging multi-step intents.

% approach
The proposed approach, inspired by the Belief-Desire-Intention model \cite{rao1995bdi}, consists of a memory module and a planning module.
The memory module enables the agent to continuously update its belief about unknown game dynamics by integrating new information inferred from the partner’s histories.
The planning module allows the agent to devise optimal actions by considering its current belief, prior knowledge, and any communicated multi-step intent.
This approach exemplifies zero-shot online planning, requiring no prior training or data, yet it ensures adaptability to accumulating environment information and strategic interaction with the partner.

% memory: weighted Beta-Bernoulli belief update
For the memory module, prior work has relied on belief updates based on natural language communication, which can introduce errors caused by human subjectivity or large language model parsing inaccuracies \cite{chen2024human}.
In contrast, our work uses a probabilistic belief model that performs Bayesian updates based on more objective information, like the partner's transition histories. The key idea is that for belief updates where accuracy matters, the agent should rely on what the partner \emph{did}, not what they \emph{said}.
Additionally, we allow the Bayesian update to be weighted by the agent's confidence in both positive and negative evidence, ensuring a more accurate belief revision process.

% planning: intent-aware MCTS via reward augmentation
For the planning module, we propose \textsc{IntentMCTS}, an intent-aware algorithm based on multi-action Monte Carlo tree search (MCTS), which allows players to take multiple actions per turn.
% how we augment reward through intents
% This algorithm integrates multi-step intent by augmenting environment reward over states included in the intent with some additional bonuses.
\textsc{IntentMCTS} integrates multi-step intent by augmenting the environment reward with a bonus when a transition lands in a desired state included in the intent.
This approach directly influences the search tree statistics, making it more natural than using intent for tie-breaking as done in \cite{chen2024human}.
% effect of reward augmentation
In addition, reward augmentation helps densify the typically sparse reward structure in coordination tasks. Unlike reward shaping based on domain knowledge \cite{dorigo1994robot,randlov1998learning} or potential functions \cite{ng1999policy}, our approach relies on time-sensitive information gathered from the partner during the interaction.
% , which is only considered in the immediate next turn (and would not affect future...).
% ablation
We evaluate four bonus schemes and find that a \textit{discounted} reward bonus leads to the most effective coordination, measured by average steps and control switches taken.

% experiment
We evaluate our approach using the Gnomes at Night testbed \cite{chen2024human}, where two players coordinate to move a single token from an initial position through a maze to a goal position. Each player sees a different maze layout and can only move the token along their visible routes. We modify the testbed so both players always see the goal, focusing on coordination rather than goal communication. We test all possible configurations across three mazes, running $10$ trials per configuration, to compare \textsc{IntentMCTS} with a shortest-path-based heuristic controller, no-intent MCTS, and single-step intent MCTS.
% results
Agent-agent simulation results show that \textsc{IntentMCTS} outperforms the other MCTS methods, taking fewer steps and control switches, validating that reward augmentation via partner intent improves coordination. The heuristic controller performs well in simple cases but struggles in more complex ones due to its lack of adaptability, achieving an overall success rate of $88\%$, while MCTS-based agents succeed over $99\%$ of the time.
% user study
Finally, we conduct a human-agent user study where each participant plays with three agents using different planning algorithms: heuristic, single-step intent MCTS, and \textsc{IntentMCTS}. 
Gameplay with the \textsc{IntentMCTS} agent achieves an 18.52\% higher success rate than the heuristic baseline and a 5.56\% improvement over single-step MCTS, along with fewer median steps, fewer control switches, and smaller interquartile ranges for both metrics.
Participants also report lower cognitive load, frustration, and higher satisfaction when partnered with the \textsc{IntentMCTS} agent compared to the other two agents.
\section{Intent-aware Coordination Problem} \label{sec:model}
We extend the shared-control game introduced in \cite{chen2024human} by allowing both players to take multiple actions per turn. We formally define the game as follows:

\begin{definition}\label{def:game}
    A \emph{multi-action shared-control game} is played between player $\ego$ and player $\human$, defined by a tuple \((\Xcal, \Scal, \Acal, \Tcal^\ego, \Tcal^\human, \Rcal)\).
    \(\Xcal\) is a finite set of environment states, with the initial state \(x_{\mathrm{init}}\) and goal state \(x_{\mathrm{goal}}\) known to both players. \(\Scal=\Xcal\times \{\ego, \human\}\) is the controller state space, consisting of tuples $(x,c)$ where $x\in \Xcal$ is the environment state and $c\in \{\ego, \human\}$ indicates which player is currently in control. The notation \(-c\) denotes the other player without control. \(\Acal\) is a finite set of actions available to both players including an action for switching control to the other player. \(\Tcal^i:\Scal\times \Acal\to\Scal\) is the deterministic transition function for player $i\in \{\ego, \human\}$. \(\Rcal^{\mathrm{env}}:\Scal\to\mathbb{R}\) is a function that assigns real-valued environment rewards to both players.
    % \begin{itemize}
    %     \item \(\Xcal\) is a finite set of environment states, with the initial state \(x_{\mathrm{init}}\) and goal state \(x_{\mathrm{goal}}\) known to both players.
    %     \item \(\Scal=\Xcal\times \{\ego, \human\}\) is the controller state space, consisting of tuples $(x,c)$ where $x\in \Xcal$ is the environment state and $c\in \{\ego, \human\}$ indicates which player is currently in control. The notation \(-c\) denotes the other player without control.
    %     \item \(\Acal\) is a finite set of actions available to both players including an action for switching control to the other player.
    %     \item \(\Tcal^i:\Scal\times \Acal\to\Scal\) is the deterministic transition function for player $i\in \{\ego, \human\}$.
    %     \item \(\Rcal^{\mathrm{env}}:\Scal\to\mathbb{R}\) is a function that assigns real-valued environment rewards to both players.
    % \end{itemize}
\end{definition}

At each timestep $t$,  when player $c$ is in control, we define the \textit{history} of the other player $-c$, denoted as $h_t^{-c} \in(\Xcal\times\Acal)^+$, as the sequence of state-action pairs taken by the other player up to timestep $t$.
We assume that at timestep $t$, player $i$ can communicate a \textit{multi-step intent}, $\zeta_t^i\in\Xcal^+$, representing a variable-length sequence of states that the player wishes the other player to visit next.

In this game, we aim to compute cooperative policies for the autonomous agent, $\ego$, when partnering with a human player, $\human$. We formalize this problem as follows:

\begin{problem}\label{problem:intent}
Given a human player whose behavior is denoted as $\pi^\human$, the task is to compute the ego player's policy $\pi^\ego$ that maximizes the expected total discounted reward:
\begin{subequations}
    \begin{flalign}
        \max_{\pi^\ego} \quad
        & \mathop{\mathbb{E}}_{a_0, a_1 \cdots} \left[ \sum_{t=0}^\infty \gamma^t \Rcal^{\mathrm{env}}(s_{t+1})\right] \label{problem:obj}\\
        \subjectto \quad
        & s_{t+1}=\Tcal^{c_t}(s_t,a_t) \\
        & a_t \sim \pi^{c_t}(s_t, h_t^{-c_t}, \zeta_t^{-c_t})
    \end{flalign}
\end{subequations} 
where $s_0=s_{\text{init}}$ and $\zeta_t^c = \mathsf{Intent Model}^c(s_t, h_t^{-c})$.
\end{problem}
\section{Related Works}\label{sec:related}

\paragraph{Intents for better coordination.}
The use of intent for improving coordination between agents can generally be categorized into two approaches: implicit \emph{inference} or explicit \emph{communication}. 
% - by inference: theory of mind, opponent modeling
Intent \emph{inference}, often explored through theory of mind \cite{byom2013theory, yoshida2008game, narang2019inferring} and opponent modeling \cite{he2016opponent, smyrnakis2010dynamic}, allows agents to predict the goals and actions of others by modeling their beliefs, desires, and intentions. 
% These methods rely on the agent's ability to infer hidden states or strategies of others. However, inference-based methods suffer from information loss due to incomplete or noisy observations of others' actions, often leading to inaccurate intent models and coordination inefficiencies.
However, these methods rely on inferring hidden states from noisy or incomplete observations, often resulting in inaccurate intent models and coordination inefficiencies.
% - by communication: imagined trajectories in CTDE fashion, unlike our approach is zero-shot online planning
Intent \emph{communication}, on the other hand, involves agents sharing explicit messages, such as future trajectories, to align their actions. Recent work has explored imagined trajectories \cite{kim2020communication}, where agents share plans during training to improve coordination. Our method, however, focuses on zero-shot, real-time planning, where agents must coordinate without prior training. Our approach also differs by generating intent as a request for \textit{what the sender player desires the other to do}, rather than predicting \textit{what the other will do}.
% Despite this progress, only a few approaches focus on making communication semantically meaningful in relation to the task at hand, which is crucial for effective human-agent coordination. Our approach is inspired partly by the BDI framework that continuously update belief of the env and model intent as desires in multi-step fashion.
Despite progress in intent communication in multi-agent reinforcement learning, few approaches ensure that the communicated intent is semantically meaningful with respect to the task at hand, which is essential for effective human-agent coordination. Our approach, inspired by the Belief-Desire-Intention model \cite{bratman1987intention, rao1995bdi}, continuously updates the agent's beliefs based on new observations made in the environment and models intents in a multi-step fashion. Building on recent work that uses single-step intents with non-probabilistic belief updates \cite{chen2024human}, our work allows multi-step intents and probabilistic belief modeling, supporting more dynamic and adaptive coordination in tasks over a longer horizon.

\paragraph{MCTS-based planning techniques.}
Monte Carlo tree search (MCTS) is an algorithm that combines tree-based search with Monte Carlo random sampling to explore and evaluate vast decision spaces efficiently \cite{browne2012survey} and has proven effective in strategic games \cite{campbell2002deep,silver2018general,silver2016mastering,silver2017mastering}, making it ideal as the foundation of our planning module.
Standard MCTS works best for turn-based games with perfect information.
% partial observability
% - for POMDP: POMCP \cite{silver2010monte}
% - for imperfect-information games: ISMCTS \cite{cowling2012information}
Several extensions have been developed to handle incomplete information.
For example, \textit{partially observable Monte Carlo planning} is designed to tackle environments modeled as \textit{partially observable Markov decision processes} using particle filtering for state estimation \cite{silver2010monte}. \textit{Information-set MCTS} tackles imperfect information games by maintaining perspective-based trees for each player, whose nodes correspond to players' information sets and edges correspond to moves from that player's viewpoint \cite{cowling2012information}. However, these methods only handle partial observations of states, not transitions as we do in our problem.
% multi-action
To extend beyond traditional turn-based games, previous research \cite{justesen2017playing, baier2018evolutionary, pipan2021application} has introduced methods for handling \textit{multi-action games}. As summarized by \cite{fujiki2015platform}, tree search in these games can either create nodes for each individual action or for the entire action sequence within a turn.
In this paper, we adopt the former approach for the benefit of a lower branching factor.
\section{Method}\label{sec:method}
This section presents an approach for an autonomous agent, acting as the ego player, to solve \Cref{problem:intent}. The agent’s decision-making process is compartmentalized into a \textit{memory module} (see \Cref{subsec:memory}) and a \textit{planning module} (see \Cref{subsec:planning}).
% The \textit{memory module} is responsible for maintaining a probabilistic belief over the unknown transition functions in the environment. It continuously updates this belief by processing the observed behaviors of the other player, using a Bayesian update mechanism to refine its understanding based on the observed history (detailed in \Cref{subsec:memory}).
% The \textit{planning module} takes in the current state and the multi-step intent communicated by its human partner to compute the most appropriate action for the agent to take, leveraging the current belief from the memory module (detailed in \Cref{subsec:planning}).

\subsection{Memory Module: Probabilistic Belief over Unknown Dynamics}\label{subsec:memory}
% why need for a memory module: tie back to the challenge of incomplete information
In \Cref{problem:intent}, the agent player $\ego$ must coordinate effectively without direct knowledge of its human partner's dynamics, $\Tcal^\human$. 
The memory module maintains a probabilistic belief over these unknown dynamics, updating it throughout the game via Bayesian inference based on the partner's transition history.
We hypothesize that the memory module will improve coordination performance as the game progresses. 
% This gives the agent an advantage over its human partners, as its memory remains consistent and does not fade as long as the updates are correct.

% how belief is modeled in the memory module: Bernoulli distribution for the presence/absence of each wall in the maze
In the game defined in \Cref{def:game} with discrete state space $\Xcal$ and action space $\Acal$, the presence of each deterministic state-action transition is modeled as a binary random variable $y\in \{0,1\}$. The belief over these transitions is represented by a Bernoulli distribution $b: \Xcal \times \Acal \to [0,1]$, indicating the likelihood of a transition’s existence. 
The probability mass function $f$ is given by: \begin{equation}\label{eqn:bernoulli}
f(y|\theta) = \theta^y (1-\theta)^{1-y}, \;\; \theta = b(x,a),
\end{equation} 
where $y=1$ when $\Tcal^\human(x,a)$ is defined and $y=0$ otherwise.

\begin{algorithm}[t]
\caption{Memory Module: Belief Update}
\label{algo:belief}
\begin{algorithmic}[1]
\State \textbf{Input:} action space $\Acal$, histories $h^\human=\{(x,a), (x,a), \dots\}$
\State \textbf{Memory:} $\alpha, \beta:\Xcal\times\Acal\to\mathbb{R}^+$, initialized to $1$ $\forall x\in\Xcal, a\in\Acal$.
\For{each $(x, a)$ in $h^\human$}
    \For{each $a'\in\Acal$}
        \If{$a' = a$}
            \State $\alpha(x,a) \gets \alpha(x,a) + c^+$
        \Else
            \State $\beta(x,a') \gets \beta(x,a') +  c^-$
        \EndIf
        \State update belief $b(x,a') \gets \frac{\alpha(x,a')}{\alpha(x,a')+\beta(x,a')}$
    \EndFor
\EndFor
\State \Return updated belief $b:\Xcal\times\Acal\to[0,1]$, 
\end{algorithmic}
\end{algorithm}

% Bayesian update mechanism given 
This belief is initialized uniformly and updated dynamically during gameplay as new evidence is gathered from the partner's movement history $h^\human$, indirectly revealing the presence or absence of transitions.
% - incoming evidence in the form of partner's histories (instead of communication information)
Unlike prior work that relies on non-probabilistic updates based on potentially erroneous communication, this approach updates based on what the partner \emph{did} over what it \emph{said}.
% - weighted by confidence in positive/negative evidence
The belief update also incorporates positive and negative evidence with different confidence factors $c^+$ and $c^-$, respectively. We ensure $c^+ > c^-$ with the intuition that the presence of a transition in the history (positive evidence) confirms its existence, but the absence of a transition (negative evidence) does not necessarily indicate its nonexistence---it may simply reflect the partner's lack of attempt.

% theorem 1: derive the posterior update rule and the constraint on weight parameters
% algo 1: summarize the belief update
\Cref{thm:weighted_bernoulli_posterior} derives the Bayesian update rule and confidence factor constraint. \Cref{algo:belief} summarizes the belief update procedure.

\begin{theorem}
% [Posterior Belief of Weighted Bernoulli Distribution]
\label{thm:weighted_bernoulli_posterior}
Let the prior belief about $ \theta = b(s,a) $ follow a Beta distribution, $ \theta \sim \text{Beta}(\alpha, \beta) $, i.e.,
\begin{equation}\label{eqn:beta}
    f(\theta \mid \alpha, \beta) = \theta^{\alpha - 1} (1 - \theta)^{\beta - 1}.
\end{equation}
We use a weighted likelihood for positive ($y = 1$) and negative ($y = 0$) evidence, with confidence factors $c^+, c^- \in \mathbb{R}^+$, where $c^+ > c^-$:
\begin{equation}\label{eqn:weighted_bernoulli}
    f(y \mid \theta) = \theta^{c^+ y} (1 - \theta)^{c^- (1 - y)}.
\end{equation}
The confidence factors must satisfy
\begin{equation}\label{eqn:weight_condition}
    c^+ = \frac{\log\left(1 - (1 - \theta)^{c^-}\right)}{\log(\theta)}.
\end{equation}
Upon observing new evidence $ y $, the posterior expectation of $ \theta $ is:
\begin{equation}\label{eqn:posterior}
    \mathbb{E}(\theta \mid y) = \frac{\alpha + c^+ y}{\alpha + c^+ y + \beta + c^- (1 - y)}.
\end{equation}
\end{theorem}

\begin{algorithm*}[t]
\caption{Planning Module: \textsc{IntentMCTS} for player $\ego$}
\label{algo:intent}
\begin{algorithmic}
    \State \textbf{Input:} current state $s_t$, intent-augmented reward $\Rcal(s_t, \zeta_t^\human)$, forward dynamics $g(s, a)$ defined in~\eqref{eqn:forward_dynamics}, feasible action set $U(s)$ defined in~\eqref{eqn:possible_actions}
    \State \textbf{Parameters:} maximum number of iterations $n=100$, exploration constant $k=\sqrt{2}$, planning discount factor $\gamma=0.99$, horizon $T=100$
    % \State \textbf{Output:} best action $a^\star$
\end{algorithmic}
\begin{algorithmic}[1]
\vspace*{-0.4cm}  % Reduce top margin before multicols
\begin{multicols}{2}
% \Procedure{IntentMCTS}{$s, \zeta^\human$}
\State Create root node $v_0$ with state $s_t$
\For{$n$ iterations}
    \State Set $v_l\gets v_0$
    \While{$v_l$ is not terminal}
    \If{$v_l$ is fully expanded}
        % get best child
        \State $v_l\gets \argmax{v'\in \mathbb{C}(v_l)}{\left(\frac{Q(v')}{N(v')}+k\sqrt{\frac{\log N(v_l)}{N(v')}}\right)}$
    \Else
        % expand
        \State $v_l\gets \textsc{Expand}(v_l)$
    \EndIf
\EndWhile
    \State $q\gets \textsc{Rollout}(s(v_l))$
    \State $\textsc{BackProp}(v_l, q)$
\EndFor
\State \Return action of best child $c^\star = \argmax{c\in \mathbb{C}(v_0)}{N(c)}$

\Statex\vspace{-2.75mm}

% \EndProcedure
\Procedure{Expand}{$v$}
    % \State Choose untried action $a$ from possible actions $A(s(v))$ wrt $\Tcal^c$ or $b(\Tcal^{-c})$
    \State Choose untried action $a$ from possible actions $U(s(v))$
    \If{$a$ is switch turn}
        % \State Create node $v'$ with $s(v')=(x(v),-c(v)), \delta(v')=1, r(v')=-1$
        \State $x,c \gets s(v)$
        \State $s'=(x,-c),\;\; \delta'=1, \;\; r'=-1$
    \Else
        % \State Create node $v'$ with $s(v'), \delta(v') = f(s(v),a)$ and $r(v') = \Rcal(s(v'), \zeta_t^\human)$
        \State $s',\delta'\gets g(s(v),a), \;\; r'\gets\Rcal(s(v'), \zeta_t^\human)$
    \EndIf
    \State Create node $v'$ with $s', \delta', r'$
    \State $\mathbb{C}(v)\gets \mathbb{C}(v)\cup\{v'\}$
\State \Return $v'$
\EndProcedure

\columnbreak

\Statex

\Procedure{Rollout}{$s$}
% rollout_from_state
\State Initialize $q=0$
\While{depth $d<T$ and $s$ not terminal}
    \State Choose $a\in U(s)$ uniformly at random
    \State $s', \delta \gets g(s, a)$
    \If{$\text{uniform-rng}(0,1) < \delta$}
        \State $s\gets s'$
    \EndIf
    \State $q\gets q + \gamma^d \Rcal(s,\zeta_t^\human)$
    \State $d\gets d+1$
\EndWhile
\State \Return $q$
\EndProcedure

\Statex

\Procedure{Backprop}{$v$, $q$}
% backpropagate
\State Initialize sample return $q_{\text{sample}} = q$ and $\delta=1$
\While{$v$ is not null}
    \State Current value estimate $w = \frac{Q(v)}{N(v)}$ if $N(v)>0$ else $0$
    \State $q_{\text{sample}} \gets r(v) + \gamma\left[\delta q_{\text{sample}} + (1-\delta) w\right]$
    \State $Q(v)\gets Q(v) + q_{\text{sample}}$ 
    \State $N(v)\gets N(v)+1$
    \State $\delta\gets \delta(v)$
    \State $v\gets$ parent of $v$
\EndWhile
\State \Return 
\EndProcedure
\end{multicols}
\vspace*{-0.4cm}  % Reduce top margin before multicols
\end{algorithmic}
\end{algorithm*}

\begin{proof}
% Weighted Likelihood and Normalization
The weighted Bernoulli likelihood is defined as $f(y \mid \theta) = \theta^{c^+ y} (1 - \theta)^{c^- (1 - y)}$ for $y \in \{0,1\}$. To ensure $ f(y \mid \theta) $ is a valid probability distribution, it must satisfy $\sum_{y\in\{0,1\}}f(y\mid\theta)=1$, thus $(1 - \theta)^{c^-} + \theta^{c^+} = 1$. Solving for $c^+$ gives \cref{eqn:weight_condition} as desired.

Applying Bayes' theorem, the posterior distribution of $\theta$ given $y$ is proportional to the product of the likelihood and the prior, i.e., 
\begin{align*}
    f(\theta | x, \alpha, \beta) &\propto f(y \mid \theta) \cdot f(\theta \mid \alpha, \beta) \\
    &= \theta^{\alpha + c^+ x - 1} (1 - \theta)^{\beta + c^- (1 - x) - 1},
\end{align*}
by substituting in the weighted likelihood from \cref{eqn:weighted_bernoulli} and the Beta prior from \cref{eqn:beta}.
% , we get $f(\theta \mid y, \alpha, \beta) \propto \theta^{\alpha + c^+ y - 1} (1 - \theta)^{\beta + c^- (1 - y) - 1}$. 
This is the kernel of a Beta distribution with updated parameters, $\text{Beta}(\alpha + c^+ y, \beta + c^- (1 - y))$. The posterior belief is therefore given by \cref{eqn:posterior} as desired.
\end{proof}

\subsection{Planning Module: Multi-Step Intent as Reward Augmentation}
\label{subsec:planning}
The planning module is the core decision-making component of the autonomous agent. We introduce an online planning algorithm called \textsc{IntentMCTS}, which considers the current state, the multi-step intent communicated by the partner, and the current belief from its memory module to compute the most appropriate action.
% base algo: multi-action MCTS
To accommodate the dynamics of players taking multiple actions per turn, this algorithm builds on multi-action Monte Carlo tree search (MCTS) \cite{justesen2017playing, baier2018evolutionary, pipan2021application}, growing trees with nodes representing each atomic action within a turn.
% Each player can take a sequence of actions in their turn, making the dynamics more complex than standard turn-based games. To handle this, we adapt Monte Carlo tree search (MCTS) to grow trees with nodes representing each atomic action within a turn, following the approach of prior work on multi-action MCTS \cite{justesen2017playing, baier2018evolutionary, pipan2021application}.
% asymmetric action selection strategy
Additionally, \textsc{IntentMCTS} selects actions differently based on the available information at self-controlled nodes versus partner-controlled nodes, following the same idea as \cite{chen2024human} in maximizing information use and growing the tree as efficiently as possible.

% notation for algorithm
\paragraph{Notations:} We denote $v$ as a node in the search tree, with $s(v)$, $r(v)$, and $\delta(v)$ representing the state, reward, and transition feasibility at node $v$, respectively. Concretely, $\delta$ is the probabilistic belief of the action that leads to node $v$ being valid. The visit count of node $v$ is denoted by $N(v)$, while $Q(v)$ represents the total discounted return at node $v$. The set of child nodes of $v$ is represented as $\mathbb{C}(v)$.

% 4 phases
\paragraph{Algorithm.}
We present \textsc{IntentMCTS} for the ego player in \Cref{algo:intent}, which performs the following four phases iteratively:

\textbf{Selection} (lines 4-10): Starting from the root node, the agent recursively selects child nodes that maximize the Upper Confidence Bound value \cite{kocsis2006bandit} until a terminal state or an expandable node is reached. A node is considered fully expanded if child nodes corresponding to all feasible actions are present.

\textbf{Expansion} (lines 15-26): Upon reaching an expandable node, the agent expands the tree by adding new child nodes based on its perception of available actions. When player $\ego$ is in control, the true transition function is known, so only feasible actions are considered. However, when player $\human$ is in control, all actions are considered, with their feasibility $\delta$ estimated by the memory module in Section \ref{subsec:memory}. Formally, we define the feasible action set as follows:
    \begin{equation}\label{eqn:possible_actions}
        U(s) =\begin{cases}
        \{a\in\Acal | \Tcal^\ego(s,a)\text{ is defined}\} &\text{if } c=\ego\\
        \Acal &\text{if } c=\human.
    \end{cases}
    \end{equation}
    For a chosen unexplored action, we construct a child node that records this action $a$, the step reward $r'$, the resulting state $s'$, and its feasibility $\delta'$. When the agent $\ego$ is in control, $s'$ is computed by its forward dynamics $\Tcal^\ego$ with $\delta' = 1$. At a human-controlled node, $s'$ is computed from $\Tcal^+(s,a)$, a transition function that returns the next state assuming the applied action $a$ is valid at state $s$, and $\delta'$ is retrieved from the memory module:
    \begin{equation}\label{eqn:forward_dynamics}
        g(s, a) := (s', \delta') = \begin{cases}
        \Tcal^\ego(s,a), 1 &\text{if } c=\ego\\
        \Tcal^+(s,a), b^\human(s, a) &\text{if } c=\human.
    \end{cases}
    \end{equation}
    Since switching control is part of the action space, the search tree can expand multiple actions from one player or switch to another anytime, accommodating the game's multi-action dynamics.

\textbf{Simulation} (lines 27-39): A rollout from the expanded node simulates future transitions until a terminal state or a predefined depth $T$ is reached. Actions are selected randomly among $U(s)$. At an agent-controlled state, the transition follows $\Tcal^\ego$. In a human-controlled state, the agent imagines the effect of an action with its memory module. The transition is executed when a random number drawn uniformly between 0 and 1 is smaller than the feasibility belief. Otherwise, the state remains unchanged. The simulation accumulates discounted rewards into the outcome $q$ and increments the depth $d$ at each step.

\textbf{Backpropagation} (lines 40-51): The rewards obtained from the simulation phase are backpropagated up the tree, updating $Q(v)$ and $N(v)$ for all nodes along the path from the expanded node back to the root. Due to the uncertain feasibility of human-controlled nodes, the rollout return depends on the feasibility of the child node from which backpropagation comes. Suppose the child node $v'$ gets a sample return $q'_\mathrm{sample}$. Its parent node's return $q_\mathrm{sample}$ should consist of three terms: (1) the step reward $r$, (2) with probability $\delta'$, the transition is feasible, and the discounted future return is $\gamma q'_\mathrm{sample}$, (3) with probability $(1 - \delta')$, the transition is invalid (action has no effect) and the discounted future return is the discounted value estimate at the current state:
    \begin{equation}
        q_\mathrm{sample} = r + \gamma\left[\delta' q'_\mathrm{sample} + (1 - \delta') \frac{Q(v)}{N(v)}\right].
    \end{equation}

\begin{figure*}[t]
    \centering
    \includegraphics[trim={0 0 0 4.5mm}, clip, width=\textwidth]{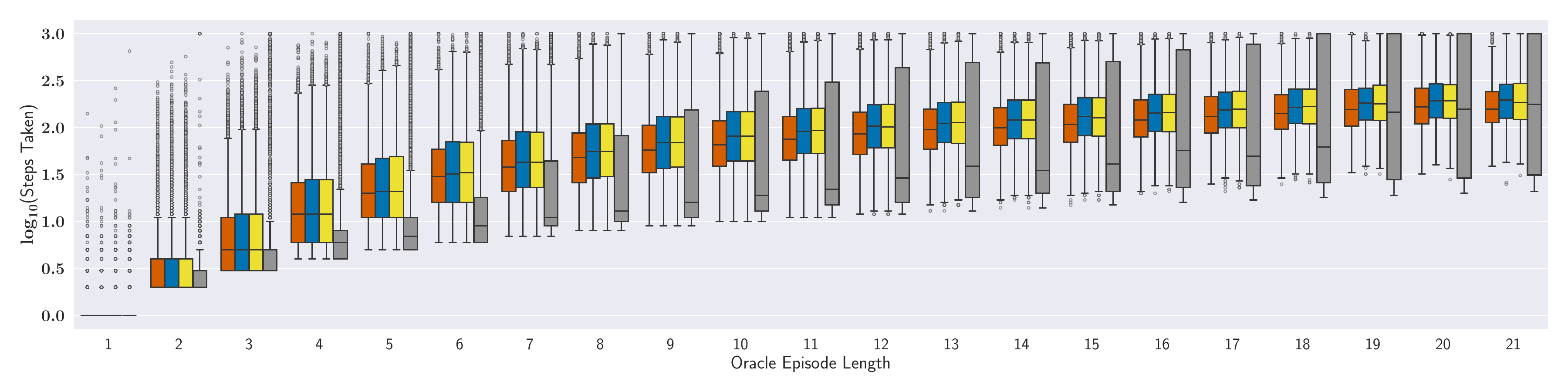}
    \includegraphics[width=\textwidth]{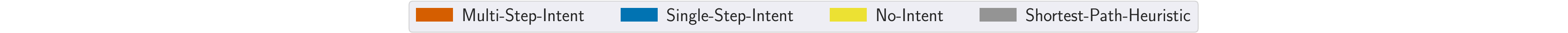}
    \includegraphics[trim={0 0 0 4.5mm}, clip, width=\textwidth]{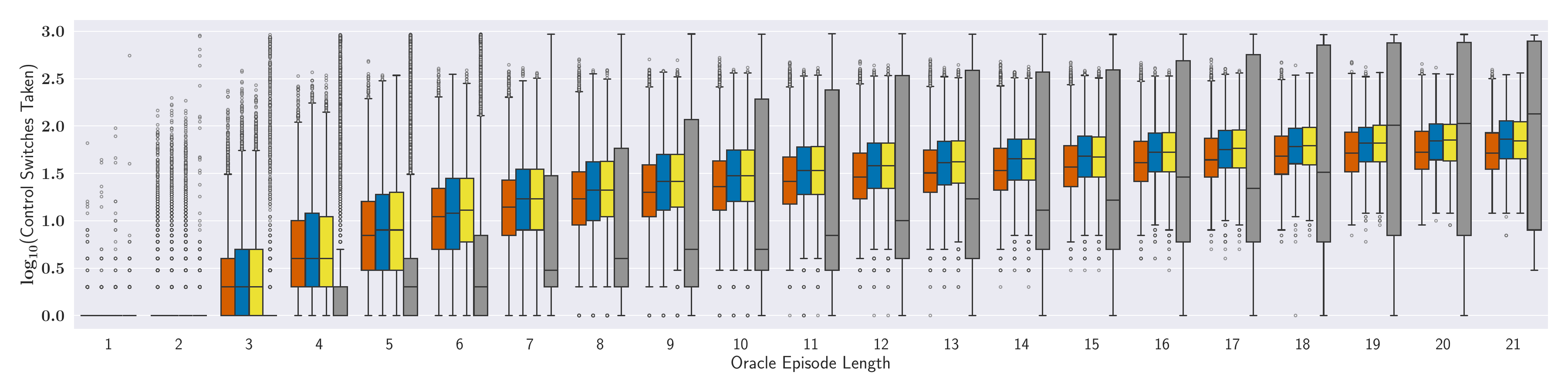}
    \vspace{-2em}
    \caption{Performance comparison of our method (orange) with three baselines, measuring steps taken (top) and control switches (bottom).
    The Y-axis is on a log-10 scale, and the X-axis represents the oracle episode length, indicating task difficulty.}
    \label{fig:all_maze_combined}
    % \vspace{-1em}
\end{figure*}

\paragraph{Reward Design.} \textsc{IntentMCTS} performs reward augmentation to encourage the agent to follow the partner's intent trajectory. We denote the environment reward function as $\Rcal^{\mathrm{env}}:\Scal\to\mathbb{R}$. Aside from the base reward, we define an intent bonus $\Rcal^\mathrm{int}:\Scal\times \Xcal^+ \to \mathbb{R}$. Since the algorithm tries to find an optimal action for player $\ego$, reward augmentation is applied only to its transitions:
\begin{equation}
    \Rcal(s, \zeta^\human) = \Rcal^{\mathrm{env}}(s) + \mathbb{I}[c=\ego]\cdot \Rcal^\mathrm{int}(s, \zeta^\human).
\end{equation}
Specifically, given an intent trajectory $\zeta^\human=\{x^\human_1, x^\human_2, \dots, x^\human_m\}$, we assign a \textit{discounted} intent bonus to provide a smooth gradient of rewards along the intent trajectory:
\begin{equation}
    \Rcal^\mathrm{int}((x,c), \zeta^\human) = \begin{cases}
        \lambda^{m-i} &\text{if } x=x^\human_i \in \zeta^\human\\
        0 &\text{otherwise.}
    \end{cases}
\end{equation}
The intent discount factor $\lambda\in(0,1)$ adaptively trades off intent trajectory following and final intent state reaching. By design, $\Rcal^\mathrm{int} \in [0, 1]$ adheres to the general principle that auxiliary objectives should not outweigh the task signals, which is a $-1$ control cost at each step in this paper.
Moreover, considering a planning discount factor $\gamma$, we can compare two extreme cases: (1) The agent follows the intent trajectory meticulously to reach the final state and (2) the agent finds a shorter path with length $n < m$ to reach the final state without visiting any intermediate intent states:
% \begin{equation}\label{eqn:bonus_case1}
%     J_\mathrm{follow} = \sum_{t = 0}^{m - 1} \gamma^t (\lambda^{m - t - 1} - 1)
%     \tag{Case 1}.
% \end{equation}
% \begin{equation}\label{eqn:bonus_case2}
%     J_\mathrm{skip} = \gamma^{n - 1} \lambda^0 + \sum_{t = 0}^{n - 1} \gamma^t (-1).
%     \tag{Case 2}
% \end{equation}
\begin{subequations}
    \begin{flalign}
        J_\mathrm{follow} &= \sum_{t = 0}^{m - 1} \gamma^t (\lambda^{m - t - 1} - 1) \tag{Case 1}.\label{eqn:bonus_case1}\\
        J_\mathrm{skip} &= \gamma^{n - 1} \lambda^0 + \sum_{t = 0}^{n - 1} \gamma^t (-1). \tag{Case 2} \label{eqn:bonus_case2}
    \end{flalign}
\end{subequations}

As shown in the equations above, the agent receives all intent bonuses in (\ref{eqn:bonus_case1}) but only receives the last intent bonus in (\ref{eqn:bonus_case2}). We analyze the benefit of skipping to the final state as follows:
\begin{equation}
    \begin{aligned}
        J_\mathrm{diff} := J_\mathrm{skip} - J_\mathrm{follow} &= \sum_{t = n - 1}^{m - 1} \gamma^t - \sum_{t = 0}^{m - 1} \gamma^t \lambda^{m - t - 1} \\
        &= \frac{\gamma^n - \gamma^{m + 1}}{\gamma (1 - \gamma)} - \frac{\gamma^m - \lambda^m}{\gamma - \lambda}.
    \end{aligned}
\end{equation}
When we choose $\lambda < \gamma$, we can compute a threshold for $n$ that makes one case more preferable:
\begin{equation}
    J_\mathrm{diff}
    \begin{cases}
        > 0 &\text{if } n < \log_\gamma{\left[\gamma^{m + 1} + \frac{\gamma (1 - \gamma) (\gamma^m - \lambda^m)}{\gamma - \lambda}\right]}\\
        \leq 0 &\text{otherwise.}
    \end{cases}
\end{equation}
As shown in the inequality above, choosing the shorter path (Case 2) is preferable to following the intent trajectory (Case 1) only for small enough $n$. Changing the hyper-parameter $\gamma$ smoothly adjusts the decision boundary between the two cases and results in different levels of intent-following behavior in \textsc{IntentMCTS}.
\section{Agent-to-Agent Simulation}
% introduce Gnomes at Night testbed
We evaluate our approach on the Gnomes at Night testbed from \cite{chen2024human}, a board game where two players coordinate to move a single token through a maze. 
The two players see different maze layouts and can only move the token along the paths within their respective mazes.
% Each player sees different maze paths, and can only move the token along their visible paths.
% config
Each game configuration $(x_{\mathrm{init}}, x_{\mathrm{goal}})$ pairs an initial and goal position, and a round is considered successful if the players reach the goal within $1000$ steps.
% modification
To focus on coordination rather than goal discovery, we modify the testbed to reveal the goal position $x_{\mathrm{goal}}$ to both players from the start, eliminating the need for communication about the goal's location and isolating the core coordination challenge.
% S, A, {T^i}, R
The state space includes all maze grid cells. Players can move right, up, left, down, or switch control. The transition function is deterministic, allowing actions unless blocked by a wall. The reward is $100$ for reaching the goal, with a penalty of $-1$ per step to encourage efficient coordination.

\subsection{Intent Model}
The agent-to-agent simulation aims to anticipate the agent's performance when playing with humans. Hence, \textsc{IntentMCTS} agents should communicate and plan with human-like intent trajectories.  
In the problem formulation, player $c$'s intent at timestep $t$, $\zeta_t^c$, is retrieved from $\mathsf{Intent Model}^c$ given the current state $s_t$ and the partner's history $h_t^{-c}$.
In our experiments, we instantiate both agents' intent models with the belief-conditioned shortest path toward the goal.
Concretely, before switching control, each agent plans the lowest-cost path as its intent, factoring in its maze layout and belief over partner transitions: taking a valid transition costs $-1$, and crossing a wall incurs a penalty of $-10(1 - \delta)$. The resulting intent minimizes steps and uncertainty: the agents prefer to traverse on their side of the maze and prioritize more feasible partner transitions when blocked.

\subsection{Comparison with Baselines}\label{subsec:baselines}
To verify the effectiveness of our \textsc{IntentMCTS} planning algorithm, we compare with three relevant baselines: a shortest-path-based heuristic controller, MCTS without intent bonus, and MCTS considering single-step intent, all using the same memory module:
\begin{itemize}[leftmargin=*]
    \item \textbf{Shortest-Path Heuristic Controller}: Same as how the intent trajectory is generated, the agent plans the lowest-cost path to the goal. If blocked by a wall, control passes to its partner. To improve the robustness of the controller, a random action is taken instead with $20\%$ chance.
    \item \textbf{No-Intent MCTS}: No reward augmentation regarding partner intent. The agents must plan with environment rewards only.
    \item \textbf{Single-Step-Intent MCTS}: Instead of reward augmentation, the intents are used as a tie-breaker during action selection. At the end of MCTS, if two child nodes are equally preferable and one follows the intent, that node is selected.
\end{itemize}

We test all configurations across three $9 \times 9$ mazes, each containing $81\times 80 = 6480$ different configurations. For each configuration, we run $10$ trials. The oracle episode length for each configuration is calculated as the minimal number of moves plus control switches needed to reach the goal, considering both players' maze layouts. This metric reflects task difficulty.
We then plot the number of steps and control switches required to complete the mazes against the oracle episode lengths for all tested trials.

\Cref{fig:all_maze_combined} shows that \textsc{IntentMCTS} consistently outperforms the other two MCTS-based methods, taking fewer steps and fewer control switches. The result validates that reward augmentation based on partner intent effectively improves coordination. The heuristic controller performs well in easy configurations. However, it lacks the flexibility to take a detour when both agents are blocked at the same grid and perform poorly in more challenging configurations. Across all game trails, the heuristic controller gets a success rate of $88\%$, whereas all MCTS-based agents succeed over $99\%$ of the time.

% table 1: ablation study among three bonus schemes
\subsection{Ablation Study}
We perform an ablation study with three alternative reward schemes to justify the discounted intent reward used in \textsc{IntentMCTS}: fixed reward, first step only reward, and length inverse reward. Note that these alternatives all range between the same bound $[0, 1]$ as $R^\mathrm{int}$, guaranteed not to outweigh the control cost of $-1$ in our setting.

% fixed
First, a \textit{fixed} intent bonus assigns a constant reward whenever the agent visits a state appearing in the intent trajectory:
\begin{equation}
    \Rcal^\mathrm{fixed}((x,c),\zeta^\human) = 0.5\cdot \mathbb{I}[x\in \zeta^\human].
\end{equation}
The fixed bonus scheme weighs all intent states equally, ignoring the temporal information within the trajectories.
% first step only
Alternatively, analogous to the simple-step MCTS, the \textit{first step only} reward only encourages visiting the immediate next intent state. Given an intent trajectory $\zeta^\human=\{x^\human_1, x^\human_2, \dots, x^\human_m\}$, we have:
\begin{equation}
    \Rcal^\mathrm{fso}((x,c),\zeta^\human) = 0.5\cdot \mathbb{I}[x = x^\human_1].
\end{equation}
% length inverse
On the other hand, the \textit{length inverse} bonus scheme prioritizes reaching the final intended state:
\begin{equation}
    \Rcal^\mathrm{linv}((x,c), \zeta^\human) = \begin{cases}
        1 &\text{if } x=x^\human_m\\
        \frac{1}{m} &\text{if } x\in\zeta^\human, x\ne x^\human_m\\
        0 &\text{otherwise.}
    \end{cases}
\end{equation}
By design, the agent receives a big bonus only when reaching the final intent state. Traversing the environment exactly as intended is unnecessary, especially for long trajectories.

As Table \ref{tab:ablation} shows, the \textit{discounted} reward bonus leads to the most effective coordination quantified by the fewest average steps and control switches taken. We report in geometric mean $\pm$ geometric standard deviation across all experiment trials because the data are positive and right-skewed.

\begin{table}[h]
    \centering 
    \begin{tabular}{lccc}
        \toprule
        \textbf{Bonus Scheme} & \textbf{Steps Taken} & \textbf{Control Switches Taken} \\
        \midrule
        fixed & $45.69 \pm 3.98$ & $17.45\pm 3.54$ \\
        first step only & $46.31 \pm 3.98$ & $18.03\pm 3.55$ \\
        length inverse & $45.98\pm 3.96$ & $18.09\pm 3.58$ \\
        \textbf{discounted} & $\mathbf{44.21\pm 3.92}$ & $\mathbf{17.11\pm3.52}$ \\
        \bottomrule
    \end{tabular} 
    \vspace{0.2em}
    \caption{Ablation study results for steps and control switches across reward bonus schemes.}
    \label{tab:ablation} 
    \vspace{-2.5em}
\end{table}
\section{Human-Agent Evaluation}
\label{sec:user_study}
We conduct a user study where human participants play with different agent partners in the Gnomes at Night testbed.

\subsection{User Study Design}
\paragraph{Independent Variables.}
The study varies the agent decision-making algorithms from the proposed algorithm and baseline methods:
\begin{description}
    \item[] \textbf{Agent Alice}: \textit{Shortest-Path-Heuristic} Controller, 
    \item[] \textbf{Agent Bob}: \textit{Multi-Step-Intent} MCTS (\textsc{IntentMCTS}), 
    \item[] \textbf{Agent Charlie}: \textit{Single-Step-Intent} MCTS. 
\end{description}
All agents share the same memory module, as detailed in \Cref{subsec:memory}. 
We exclude No-Intent MCTS, as \cite{chen2024human} has already demonstrated that single-step intent performs better than no-intent.

% objective measures: bar charts
% subjective ratings: radar chart
\begin{figure*}[ht]
    \centering
    \includegraphics[width=\linewidth]{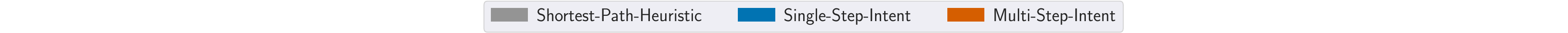}
    \begin{subfigure}[b]{0.22\linewidth}
        \centering
        \includegraphics[width=\linewidth]{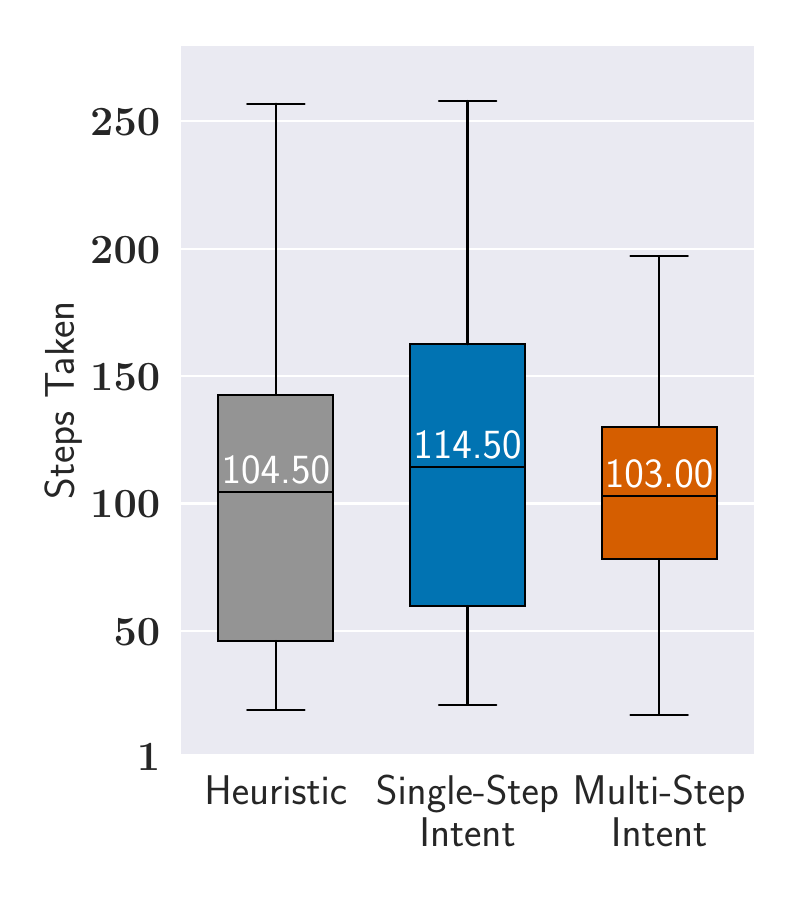}
        \caption{Steps Taken}
        \label{fig:steps}
    \end{subfigure}%
    % \hfill
    \begin{subfigure}[b]{0.22\linewidth}
        \centering
        \includegraphics[width=\linewidth]{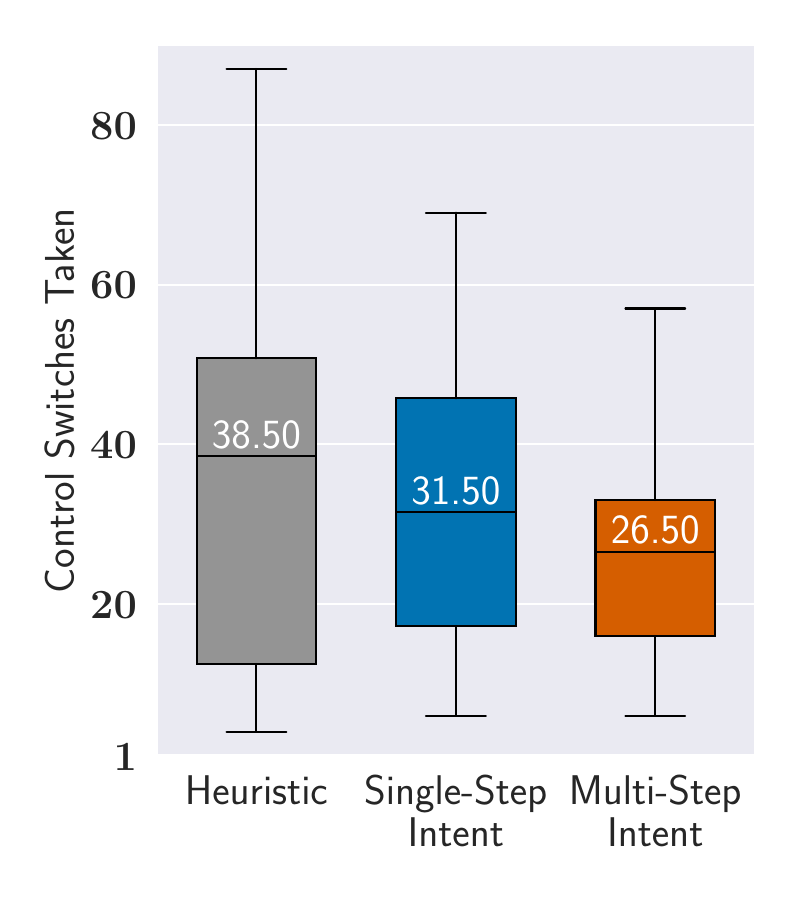}
        \caption{Control Switches Taken}
        \label{fig:switches}
    \end{subfigure}%
    % \hfill
    \begin{subfigure}[b]{0.22\linewidth}
        \centering
        \includegraphics[width=\linewidth]{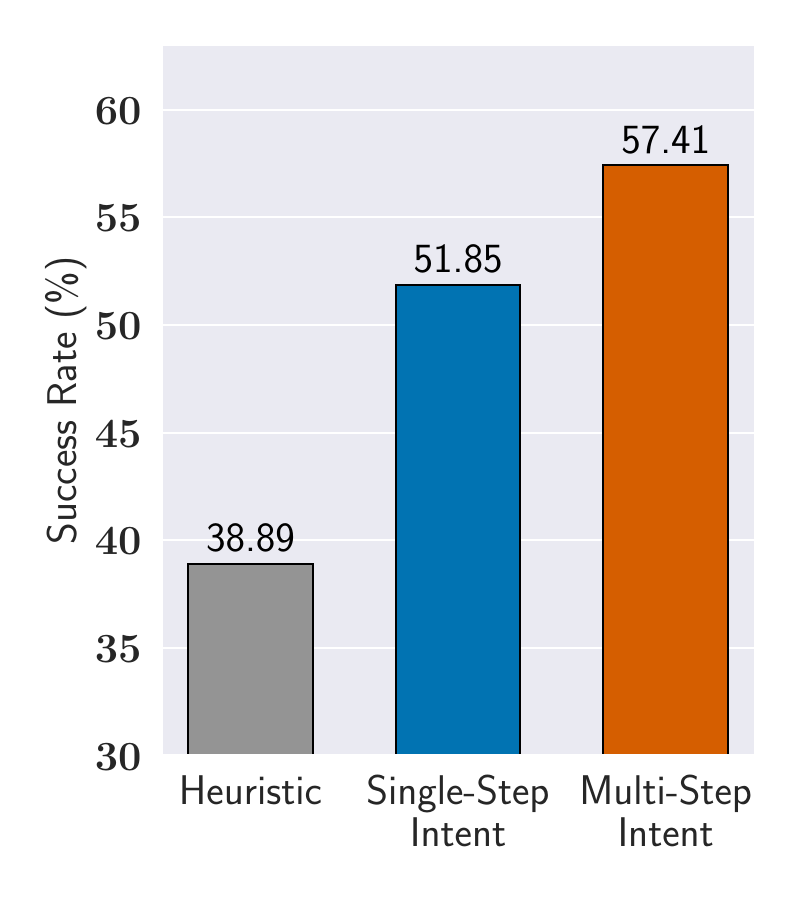}
        \caption{Success Rates}
        \label{fig:success_rate}
    \end{subfigure}%
    % \hfill
    \begin{subfigure}[b]{0.33\linewidth}
        \centering
        \includegraphics[width=\linewidth]{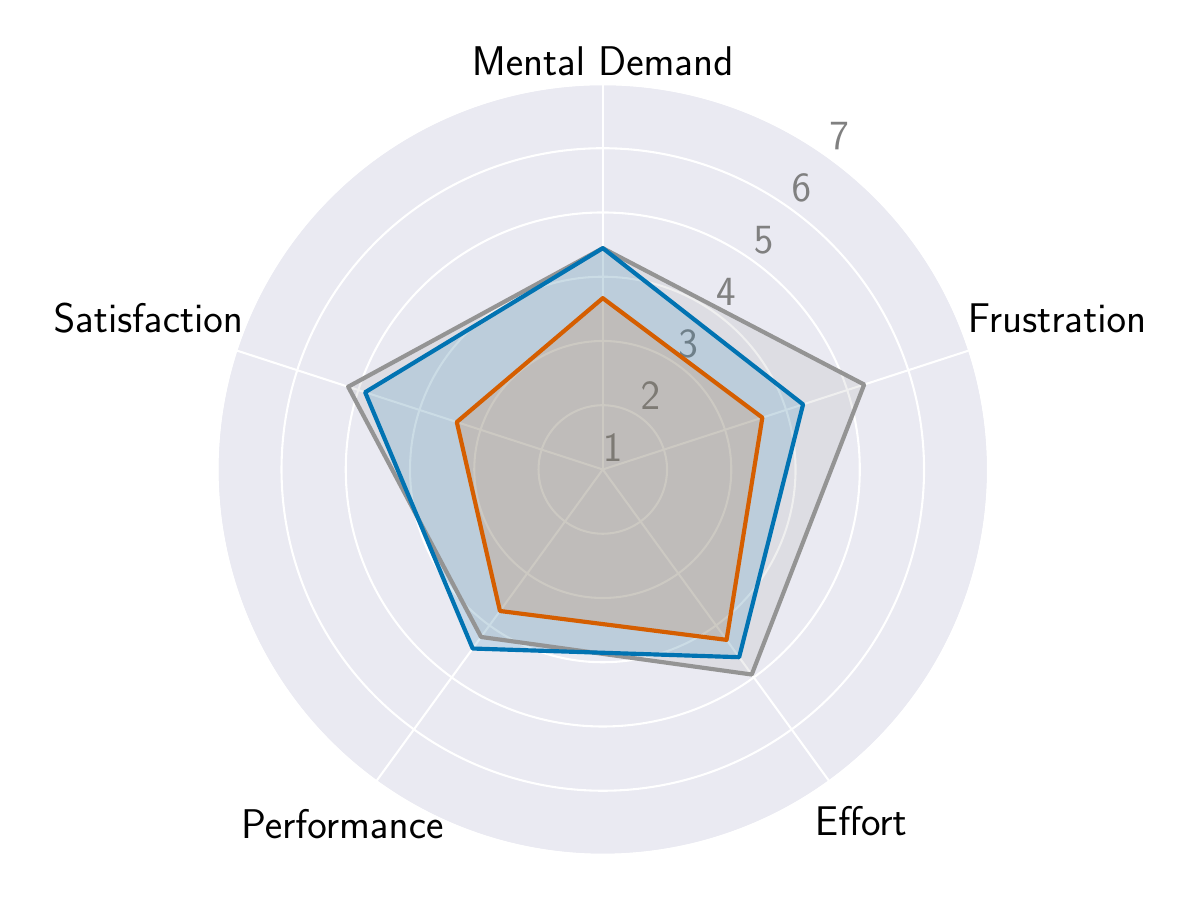}
        \caption{Survey Ratings}
        \label{fig:radar}
    \end{subfigure}
    \vspace{-0.5em}
    \caption{(a, b) Box plots for steps and control switches taken, with medians labeled in white text; (c) Bar chart for average success rates; (d) Radar chart for average ratings from 7-point Likert scale survey (Lower values are preferred in all questions, e.g., 1$=$very satisfied and 7$=$not satisfied at all).}
    \label{fig:combined}
    \vspace{-1em}
\end{figure*}

% belief progression
\begin{figure*}[ht]
    \centering
    \begin{subfigure}{0.18 \linewidth}
        \centering
        \includegraphics[width=\linewidth]{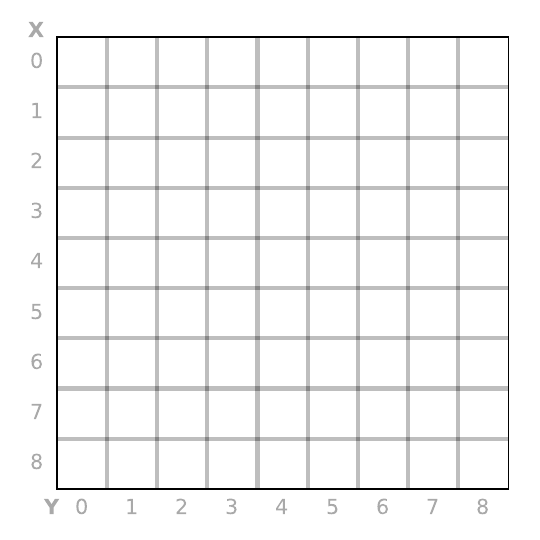}
        \caption{Step $0$}
    \end{subfigure}
    \begin{subfigure}{0.18 \linewidth}
        \centering
        \includegraphics[width=\linewidth]{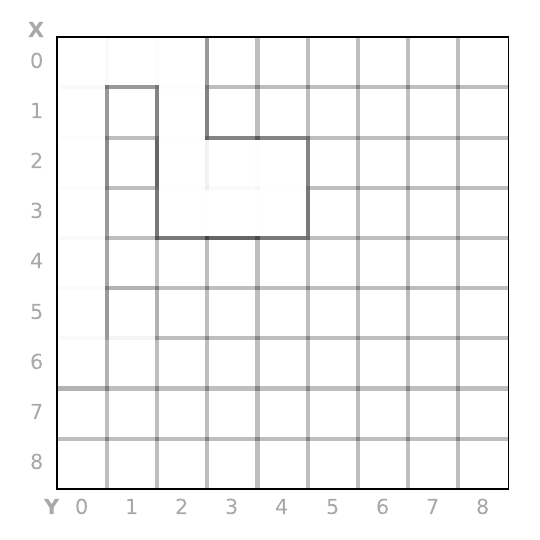}
        \caption{Step $100$}
    \end{subfigure}
    \begin{subfigure}{0.18 \linewidth}
        \centering
        \includegraphics[width=\linewidth]{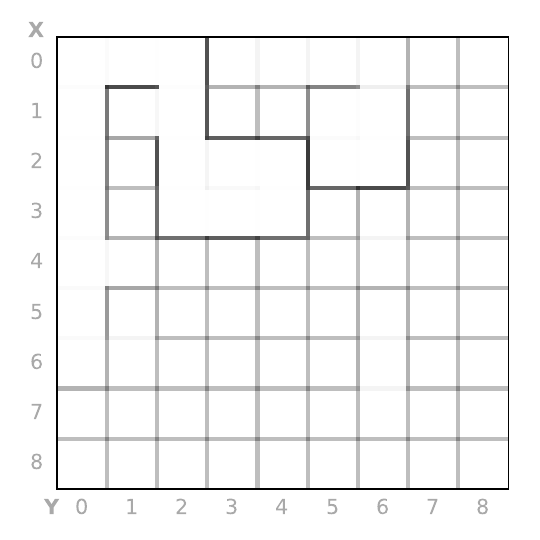}
        \caption{Step $250$}
    \end{subfigure}
    \begin{subfigure}{0.18 \linewidth}
        \centering
        \includegraphics[width=\linewidth]{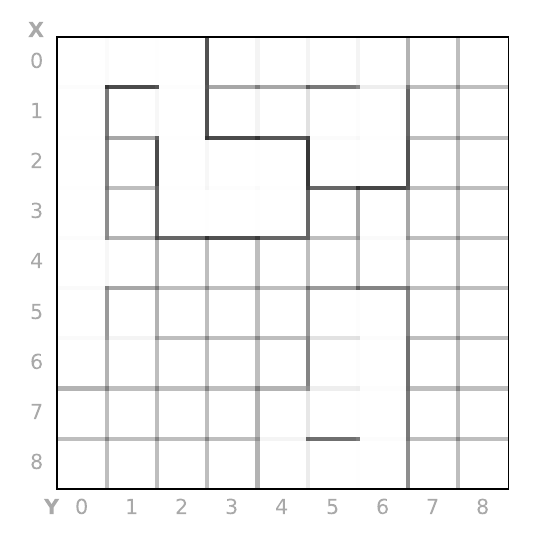}
        \caption{Step $350$}
    \end{subfigure}
    \begin{subfigure}{0.18 \linewidth}
        \centering
        \includegraphics[width=\linewidth]{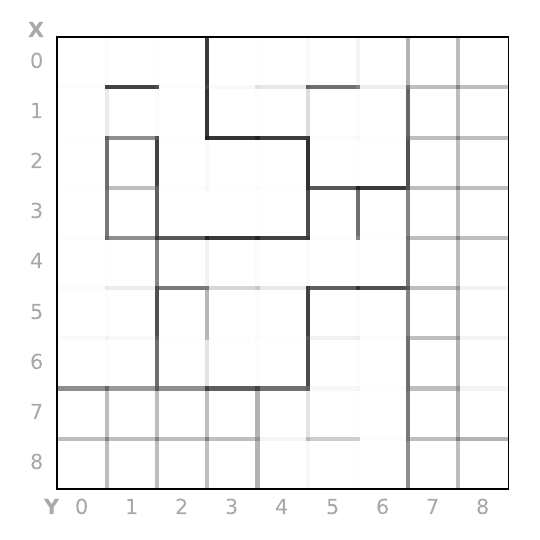}
        \caption{Step $620$}
    \end{subfigure}
    \raisebox{0.58cm}{\includegraphics[trim={9.15cm 0 0 0},clip,height=3.18cm]{figures/belief_legend.pdf}}
    \vspace{-0.5em}
    \caption{Visualization of the agent's belief of the wall layout in human’s maze at key steps in gameplay. Darker lines indicate a stronger belief in the presence of walls. 
    % The full progression is available in the supplementary video \texttt{``example\_replay.mp4."}
    }
    \label{fig:belief_progression}
    % \vspace{-1em}
\end{figure*}

\paragraph{Dependent Variables.}
We measure the number of steps taken, the number of control switches taken, and whether the token reaches the goal position. 
% We also ask participants to rate based on a 7-point Likert scale survey on the following questions.
% \begin{quote}
% \it
%     (1) How mentally demanding was the task?
%     (2) How physically demanding was the task?
%     (3) How hurried or rushed was the pace of the task?
%     (4) How insecure, discouraged, irritated, stressed, and annoyed were you?
%     (5) How hard did you have to work to accomplish your level of performance?
%     (6) How successful were you in accomplishing what you were asked to do?
%     (7) How satisfied are you with the performance of your partner?
%     \begin{enumerate}
%     % \it
%     \item How mentally demanding was the task? 
%     \item How physically demanding was the task?
%     \item How hurried or rushed was the pace of the task?
%     \item How insecure, discouraged, irritated, stressed, and annoyed were you?
%     \item How hard did you have to work to accomplish your level of performance?
%     \item How successful were you in accomplishing what you were asked to do?
%     \item How satisfied are you with the performance of your partner?
% \end{enumerate}
% \end{quote}
Participants also complete a 7-point Likert survey: 
The first six questions are from the NASA-TLX \cite{hart1988development}, covering mental demand, physical demand, temporal demand, frustration, effort, and performance, respectively. The final question assesses participant satisfaction with each agent partner.

\paragraph{Hypotheses.} We design the experiment with two hypotheses: 
\begin{description}
    \item [(H1)] Gameplay with the \textit{Multi-Step-Intent} agent will outperform the other two agents in coordination efficiency.
    \item [(H2)] Participants will report better subjective survey scores on the \textit{Multi-Step-Intent} agent than other two agents.
\end{description}

\begin{table}[t]
    \begin{tabular}{cccc}
        \toprule
        \textbf{Method} & \textbf{Session 1} & \textbf{Session 2} & \textbf{Session 3} \\ 
        \midrule
        \textit{Shortest-Path-Heuristic} & 0.37 & 0.37 & 0.26 \\ 
        \textit{Single-Step-Intent} & 0.32 & 0.26 & 0.42 \\ 
        \textit{Multi-Step-Intent} & 0.31 & 0.37 & 0.32 \\ 
        \bottomrule
    \end{tabular}
    \vspace{0.2em}
    \caption{Even distribution of methods across session orders.}
    \label{tab:counterbalancing}
    \vspace{-2.5em}
\end{table}

\paragraph{Experiment Design.} We use a \textit{within-subject design} where each participant completes three game sessions, each partners with one of the agents---Alice, Bob, or Charlie---as outlined in the independent variables. All sessions use the same maze, and each session includes the same set of three distinct configurations to ensure consistency for comparison. We ask participants to answer the 7-item survey after each session.
% randomizations
To control for order effects---such as learning, fatigue, and carryover---we use \textit{counterbalancing} to randomize the order of sessions \cite{martin2007doing}. The counterbalancing ensures an even distribution of methods across session orders, as shown in \Cref{tab:counterbalancing}.
Within each session, we also randomize the order of configurations to prevent boredom or disengagement over time. 
This two-level randomization design reduces potential biases, allowing for a more accurate comparison of participant performance across methods. 
% An example of the user study can be found in the supplementary file \texttt{``user\_study\_procedure.pdf."}

\paragraph{Participants.} The study recruits $18$ university students as consenting participants, with an average age of $26.28$. The gender distribution was $0.83$ male, $0.11$ female, and $0.06$ non-binary. 
% All participant data are anonymized and provided as CSV files in the \texttt{``user\_data"} folder in the supplementary.

\subsection{Results and Discussions}

\paragraph{Regarding (H1).}
We present the distributions of steps and control switches for each method using boxplots, with median values labeled in white text. 
In \Cref{fig:steps}, \textit{Multi-Step-Intent} shows a median step count $1.5$ steps lower than \textit{Shortest-Path-Heuristic} and $11.5$ steps lower than \textit{Single-Step-Intent}, with a significantly smaller interquartile range (IQR), indicating less variation due to randomness or participant differences.
Similarly, \Cref{fig:switches} shows \textit{Multi-Step-Intent} requires a median of $12$ fewer control switches than \textit{Shortest-Path-Heuristic} and $5$ fewer than \textit{Single-Step-Intent}, again with a much smaller IQR. 
Additionally, \Cref{fig:success_rate} highlights that \textit{Multi-Step-Intent} improves success rates by $5.56\%$ over \textit{Single-Step-Intent} and $18.52\%$ over \textit{Shortest-Path-Heuristic}.
These results support (H1), demonstrating improved coordination efficiency.

\paragraph{Regarding (H2).}
We present average participant ratings for $5$ of the $7$ survey items in \Cref{fig:radar}, excluding physical and temporal demand as they are less relevant to the task. Lower scores indicate lower cognitive load or higher satisfaction. The \textit{Multi-Step-Intent} agent achieves lower scores in all dimensions than the other two agents, shown as the smallest orange area, supporting (H2) and demonstrating an enhanced user experience.

% belief heat maze progression to show memory effect
\paragraph{Effect of Memory Module in Gameplay.}
In the Gnomes at Night testbed, players' private transition functions reflect the unique wall layouts in their mazes.
\Cref{fig:belief_progression} showcases how the agent's belief about the human's maze walls, stored in its memory module, evolves during a human-agent sample gameplay. 
Darker lines indicate a stronger belief in the presence of walls. 
Starting with a uniform belief of $0.5$ for all walls, the agent refines its understanding, identifying one room's boundaries by step $100$ and inferring the overall maze layout by the end of the game at step $620$.

\paragraph{Qualitative Observations.}
The quantitative results from both objective metrics and subjective ratings clearly indicate that \textit{Multi-Step-Intent} outperforms \textit{Shortest-Path-Heuristic} and \textit{Single-Step-Intent}. Qualitative feedback from participants reveals additional insights. Many participants appreciate the explainability provided by the multi-step intent algorithm, which enhances their satisfaction even when the performance difference is not immediately evident. We also observe that participants use alternative strategies, such as marking cells around closed rooms to signal being trapped rather than moving along the room boundary repeatedly. However, frustration arises when the agent seems to ignore their intended paths, suggesting that incorporating natural language communication or clarification as a valuable next step.
\section{Conclusion}\label{sec:conclusion}
We extend a turn-based shared-control game under incomplete information to allow multiple actions per turn, enabling the use of multi-step intent to enhance performance in long-horizon tasks. We develop \textsc{IntentMCTS}, an online planning algorithm that incorporates a memory module for probabilistic beliefs and leverages multi-step intent through reward augmentation.
Both agent-to-agent simulations and a human-agent user study show that \textsc{IntentMCTS} outperforms baseline methods in terms of steps, turns, and success rates. The user study also highlights improvements in participant cognitive load and satisfaction toward the partner.
% These findings confirm the benefits of multi-step intent in improving human-agent collaboration.

% Future work
% - intent not only for reward but also as constraint when viable
% - communication -> multi-step intent; also using communication to update belief in memory
% - intent generation better than shortest path heuristic
% - deterministic -> dynamic transition function
% For future work, we can explore using multi-step intent as a constraint when feasible, based on the observation that humans expect agents to follow their intent whenever possible. We can also extract multi-step intent from natural language communication and leverage it for more efficient belief updates or as a supplementary source of information alongside transition history. Finally, we aim to transition from a handcrafted heuristic to a data-driven model for intent generation.
For future work, we can explore using multi-step intent as a constraint when feasible, extracting it from natural language for belief updates, and transitioning from a heuristic to a data-driven intent generation model.

%%%%%%%%%%%%%%%%%%%%%%%%%%%%%%%%%%%%%%%%%%%%%%%%%%%%%%%%%%%%%%%%%%%%%%%%

%%% The acknowledgments section is defined using the "acks" environment
%%% (rather than an unnumbered section). The use of this environment 
%%% ensures the proper identification of the section in the article 
%%% metadata as well as the consistent spelling of the heading.

\begin{acks}
This work was supported by NSF CNS-1836900, NSF CPS-2133481, ARL W911NF-23-2-0011 and the Lockheed Martin Corporation. Any opinions, findings, conclusions, or recommendations expressed in this material are those of the authors and do not necessarily reflect the views of the sponsors.
\end{acks}

%%%%%%%%%%%%%%%%%%%%%%%%%%%%%%%%%%%%%%%%%%%%%%%%%%%%%%%%%%%%%%%%%%%%%%%%

%%% The next two lines define, first, the bibliography style to be 
%%% applied, and, second, the bibliography file to be used.
\balance
\bibliographystyle{ACM-Reference-Format} 
\bibliography{ref}

%%%%%%%%%%%%%%%%%%%%%%%%%%%%%%%%%%%%%%%%%%%%%%%%%%%%%%%%%%%%%%%%%%%%%%%%

\end{document}